\documentclass[10pt,twocolumn,letterpaper]{article}

\usepackage{cvpr}              
\usepackage{bm}
\usepackage{multirow}
\usepackage{color}
\usepackage{colortbl}
\usepackage{appendix}
\definecolor{grayfont}{RGB}{150,150,150}
\definecolor{mygray}{gray}{.9}

\usepackage[dvipsnames]{xcolor}

\definecolor{cvprblue}{rgb}{0.21,0.49,0.74}
\usepackage[pagebackref,breaklinks,colorlinks,citecolor=cvprblue]{hyperref}

\def\paperID{9961} 
\def\confName{CVPR}
\def\confYear{2024}

\title{How to Make Cross Encoder a Good Teacher \\ for Efficient Image-Text Retrieval?}

\author{\vspace{-2.3em} \\Yuxin Chen\textsuperscript{1,2,3*}, Zongyang Ma\textsuperscript{1,2,3*}, Ziqi Zhang\textsuperscript{1*}, Zhongang Qi\textsuperscript{2}, Chunfeng Yuan\textsuperscript{1$\dagger$}, \\Bing Li\textsuperscript{1}, Junfu Pu\textsuperscript{2}, Ying Shan\textsuperscript{2}, Xiaojuan Qi\textsuperscript{5}, Weiming Hu\textsuperscript{1,3,4}\\
\textsuperscript{1}State Key Laboratory of Multimodal Artificial Intelligence Systems,\\Institute of Automation, Chinese Academy of Sciences; 
\ \  \textsuperscript{2}ARC Lab,  Tencent PCG;\\
\textsuperscript{3}School of Artificial Intelligence, University of Chinese Academy of Sciences;\\
\textsuperscript{4}School of Information Science and Technology, ShanghaiTech University;
\ \ \textsuperscript{5}The University of Hong Kong\\
{\tt\small \{chenyuxin2019,mazongyang2020\}@ia.ac.cn, \{ziqi.zhang,cfyuan,bli,wmhu\}@nlpr.ia.ac.cn} \\
\tt\small \{zhongangqi,jevinpu,yingsshan\}@tencent.com, xjqi@eee.hku.hk
\vspace{-1.7em}
}

\begin{document}
\maketitle
\footnotetext{* Equal contribution. $\dagger$ Corresponding author.}

\begin{abstract}
Dominant dual-encoder models enable efficient image-text retrieval but suffer from limited accuracy, while the cross-encoder models offer higher accuracy at the expense of efficiency.
Distilling cross-modality matching knowledge from cross-encoder to dual-encoder provides a natural approach to harness their strengths. Thus, we investigate the following valuable question: how to make cross-encoder a good teacher for dual-encoder?
Our findings are threefold: 
(1) Cross-modal similarity score distribution of cross-encoder is more concentrated, while the result of dual-encoder is nearly normal, making vanilla logit distillation less effective. 
However, ranking distillation remains practical, as it is not affected by the score distribution.
(2) Only the relative order between hard negatives conveys valid knowledge, while the order information between easy negatives has little significance.
(3) Maintaining the coordination between distillation loss and dual-encoder training loss is beneficial for knowledge transfer.
Based on these findings, we propose a novel Contrastive Partial Ranking Distillation (CPRD) method, which implements the objective of mimicking relative order between hard negative samples with contrastive learning. This approach coordinates with the training of the dual-encoder, effectively transferring valid knowledge from the cross-encoder to the dual-encoder.
Extensive experiments on image-text retrieval and ranking tasks show that our method surpasses other distillation methods and significantly improves the accuracy of dual-encoder.
\end{abstract}

\vspace{-0.5cm}
\section{Introduction}
\vspace{-0.2cm}
\label{sec:intro}

\begin{figure}
    \centering
    \includegraphics[width=\linewidth]{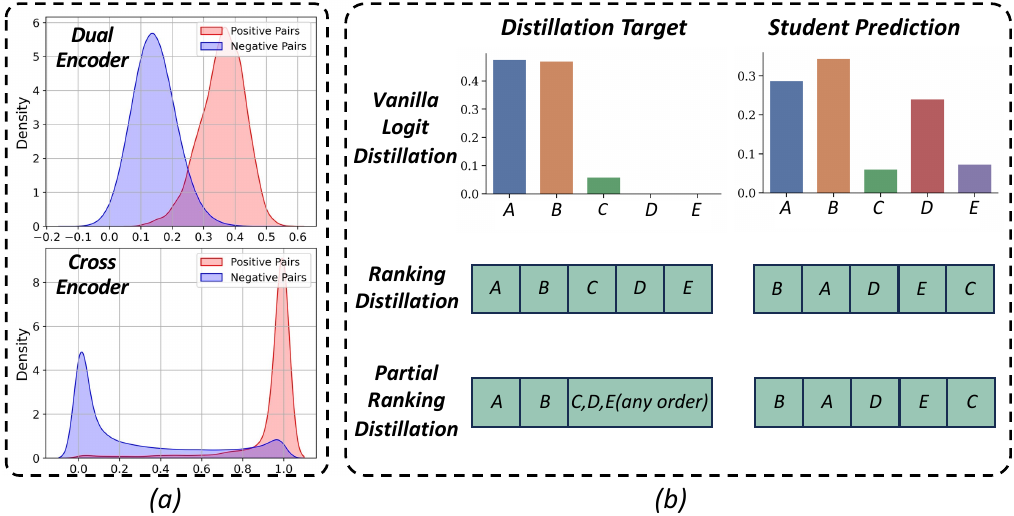}
    \vspace{-0.8cm}
    \caption{(a) Similarity score distribution of dual-encoder and cross-encoder. (b) Student predictions and targets for different types of distillation methods. For partial ranking distillation, the relative order between easy negatives is disregarded.
    }
    \label{fig:abstract}
    \vspace{-0.7cm}
\end{figure}

Dominant pre-training works for image-text retrieval adopt dual-encoder architecture \cite{radford2021learning, jia2021scaling, wen2021cookie, lu2022cots} to separately extract image and text representations and employ contrastive learning to facilitate global image-text alignment.
Although dual-encoder has high efficiency, its retrieval accuracy is sub-optimal due to the lack of modality interactions except for the final dot product. 
In contrast, the cross-encoder architecture of another line of works perform deep cross-modal interaction between images and texts in a single encoder.
This allows them to retrieve more accurately, but the need to calculate on all possible image-text pairs sacrifices retrieval efficiency.

To obtain an efficient and accurate retriever, a natural idea is to distill the cross-modal matching knowledge from the cross-encoder into the dual-encoder.
However, the discrepancies between the two models make the existing attention and logit distillation methods less versatile or effective.
On the one hand, attention distillation requires extra modules to compute cross-modal attention of dual-encoder and a cross-encoder with specific architecture to provide distillation targets.
On the other hand, the differences in training objectives and architectures between two models lead to significantly distinct and non-directly alignable cross-modal similarity score distributions. 

To explore a general and efficient way to achieve effective knowledge transfer from cross-encoder to dual-encoder, we analyze their distinct characteristics and the difficulties hindering distillation, and make the following observations:
(1) As shown in Figure \ref{fig:abstract}(a), the cross-modal similarity distribution of the dual-encoder constrained by contrastive learning tends to be "moderate" and normally distributed, whereas the similarity score distribution of cross-encoder optimized with image-text matching is more ``aggressive" and mainly concentrates on 0/1.
Vanilla logit distillation based on KL-divergence enforces the similarity score distribution of dual-encoder to approximate that of cross-encoder. It is difficult due to the substantial differences between them and may interfere with the original training objective of dual-encoder, thereby disrupting the learning of coarse image-text alignment.
However, we find that mimicking the ranking of negatives derived from sorting similarity scores is an effective distillation scheme, which we refer to as ranking distillation, as it solely depends on the relative ranking of the scores and is not influenced by the sharpness of score distribution.
(2) Furthermore, performance gap between dual-encoder and cross-encoder mainly stems from inaccurate ranking of top-ranked items. 
Supporting this claim, re-ranking top 32 items retrieved by dual-encoder with cross-encoder substantially improves retrieval accuracy. Further expanding the number of re-ranked items yields negligible gains, indicating that distillation should focus more on partial ranking of top-ranked samples rather than all samples. The illustrations of vanilla logit distillation, ranking distillation, and partial ranking distillation are shown in Figure \ref{fig:abstract}(b) for clarity.
(3) Different types of loss have varied effects on the image-text embedding space of dual-encoder. Therefore, it is beneficial to maintain coordination between ranking distillation loss and the original training loss of dual-encoder to avoid interfering with the learning process.

Based on the above observations, we propose a novel Contrastive Partial Ranking Distillation (CPRD) method to enable effective ranking knowledge distillation. Specifically, we learn the ranking of hard negative samples via contrastive learning. Given an image, we first utilize the dual-encoder to identify the top-K hard negative texts and obtain the ranking of these texts. Then, we feed the image and negative texts into cross-encoder to compute the matching scores and divide the negative texts into valid and invalid negative texts. Valid negative texts have a larger matching score with the image, and their relative order contains rich cross-modal matching knowledge. Therefore, we employ contrastive learning to pull higher-ranked valid negative texts closer to image while pushing the lower-ranked texts away, ensuring the consistency between ranking of valid negative texts from dual-encoder and that from the cross-encoder. On the other hand, invalid negative texts have a smaller matching score with the image, and their relative order does not contain valid information. Thus, we only employ contrastive learning to push all invalid negative texts away and disregard the relative order between them.

This approach does not require the similarity distributions of dual-encoders and cross-encoders to be similar, overcoming the distillation difficulties caused by the substantial differences between similarity distributions.
Furthermore, we achieve the objective of hard negative ranking learning with contrastive learning, aligning seamlessly with the dual-encoder training process.
The contributions of this work can be summarized as follows:
\begin{itemize}
    \item We conduct a comprehensive investigation into the effective knowledge distillation from cross-encoders to dual-encoders and identify three key aspects.
    \item We propose the Contrastive Partial Ranking Distillation (CPRD) method, which attains the objective of learning the relative order between valid hard negatives via contrastive learning, enabling effective knowledge transfer from cross-encoder to dual-encoder.
    \item Experimental results show that our method outperforms previous distillation methods, significantly improving the retrieval and ranking accuracy of dual-encoder.
\end{itemize}

\vspace{-0.1cm}
\section{Related Work}
\vspace{-0.1cm}
\label{sec:relatedwork}
\noindent \textbf{Pre-training for Image-text Retrieval.}
Pre-training works for image-text retrieval can be categorized into dual-encoder methods and cross-encoder methods. Dual-encoder methods \cite{radford2021learning, jia2021scaling, wen2021cookie, lu2022cots, sun2021lightningdot} adopt two separate encoders to extract the image and text features separately, and employ contrastive learning to align global representations in a shared embedding space. 
Cross-encoder methods \cite{li2019visualbert, lu2019vilbert, chen2019uniter, li2020oscar, zhang2021vinvl, li2021align, zeng2021multi, li2022blip, bao2022vlmo} employ a single encoder for the joint encoding of image-text features. Many proxy tasks are proposed to facilitate cross-modal interaction in the cross-encoder, \textit{e.g.,} masked language modeling (MLM), masked region modeling (MRM), and masked image modeling (MIM), \textit{etc.}
In order to improve the retrieval accuracy of dual-encoders, one line of approaches \cite{sun2021lightningdot, lu2022cots} draw inspiration from cross-encoder and enhance the cross-modal interaction by adapting MLM, MRM, and MIM tasks to dual-encoder. However, another approach, transferring knowledge from cross-encoder to dual-encoder through distillation, has not been fully explored.

\noindent \textbf{Knowledge Distillation from Cross-Encoder to Dual-Encoder.} 
Previous works distilling knowledge from the cross-encoder to the dual-encoder can be divided into attention distillation methods\cite{wang2022distilled} and logit distillation methods\cite{miech2021thinking}.
Attention distillation methods aim to align the cross-modal attention of two models, which requires two prerequisites: (1) Both models have to adopt attention-based backbone, \textit{e.g.,} ViT\cite{dosovitskiy2020image}, BERT\cite{kenton2019bert}, to produce attention map. (2) Input to dual-encoder and cross-encoder is exactly identical to guarantee their attention maps have the same shape and semantics. 
These make attention distillation less versatile. 
Inspired by the distillation works in image classification, Miech et al. \cite{miech2021thinking} and Lei et al. \cite{lei2022loopitr} introduce logit distillation into image-text retrieval. 
The core idea is to constrain the consistency of the image-text similarity score distribution of the dual encoder and cross-encoder through KL-divergence based loss. 
But substantial similarity distribution differences between dual-encoder and cross-encoder make it difficult to transfer knowledge effectively.
To explore an effective and general distillation method to improve the performance of dual-encoders, we investigate and identify three key factors and further propose the Contrastive Partial Ranking Distillation method.

\noindent \textbf{Ranking Distillation for Neural Ranking.} In the field of textual neural ranking, there are some studies on ranking distillation \cite{menon2022defense, reddi2021rankdistil, hofstatter2020improving}. Sashank et al.\cite{reddi2021rankdistil} propose rank distillation losses with cross-entropy or MSE loss to constrain the consistency of positive sample scores, which is less effective due to the significant differences in similarity score distribution in image-text retrieval.
Sebastian et al.\cite{hofstatter2020improving} 
propose Margin-MSE, which requires the same margin between positive and negative sample scores for student and teacher models. Aditya et al.\cite{menon2022defense} further propose M3SE, 
requiring the student and teacher models to have the same margin between positive and the hardest negative sample.
However, it only considers the hardest negative samples, limiting the knowledge that can be transferred. Moreover, the MSE loss is not coordinated with the contrastive learning for dual-encoder training, resulting in interference in the learning process. In contrast, our proposed CPRD method considers the relative order among multiple hard negative samples via contrastive learning, aligning with the original training loss of dual-encoder.

\begin{figure*}
    \centering
    \includegraphics[width=\textwidth]{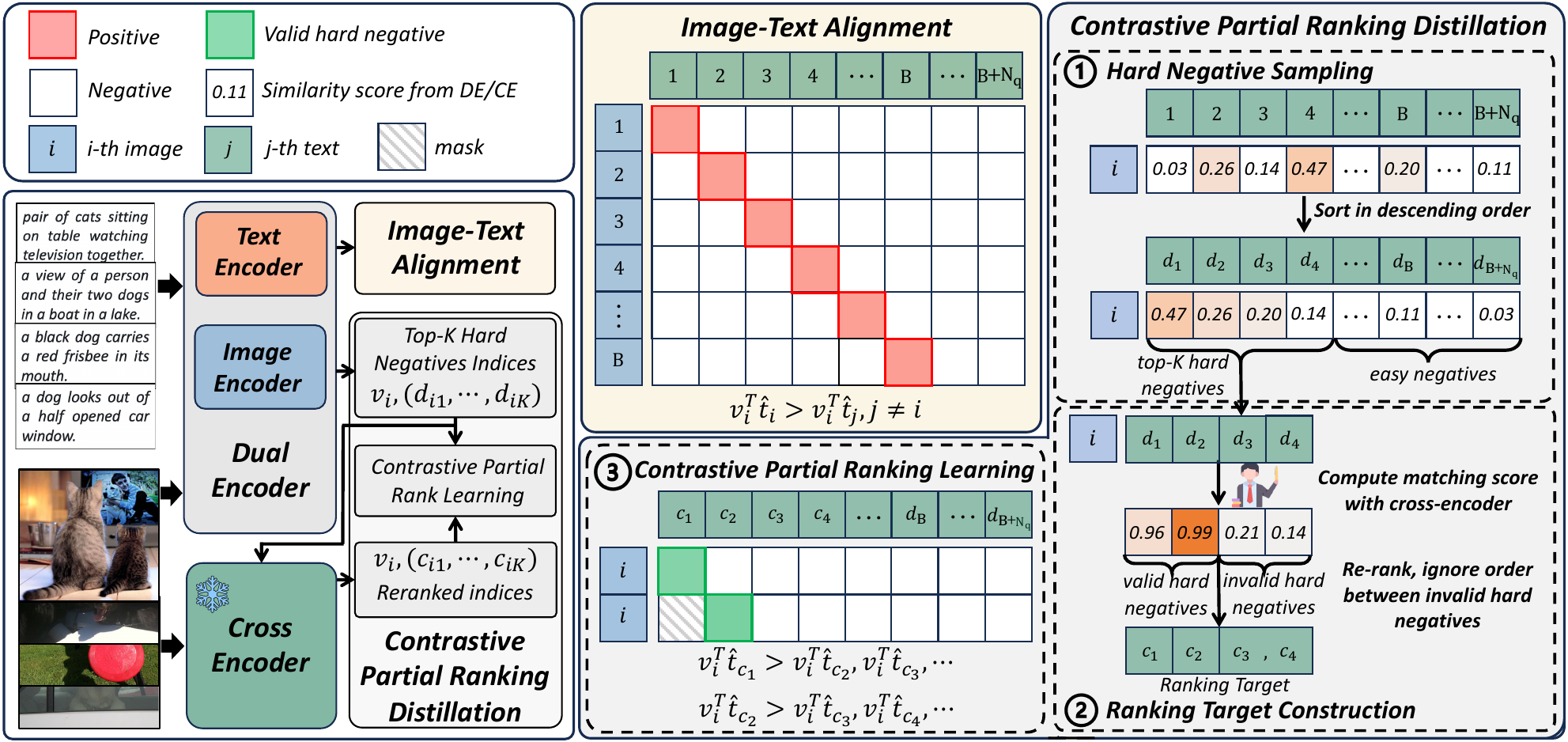}
    \vspace{-0.7cm}
    \caption{The illustration of the Contrastive Partial Ranking Distillation method. The left shows the overall training process and the right side elaborates on the computation process of image-text alignment and contrastive partial ranking distillation.}
    \label{fig:framework}
    \vspace{-0.6cm}
\end{figure*}

\vspace{-0.1cm}
\section{Method}
\vspace{-0.1cm}

In this section, we first introduce the architecture and training objective of vanilla dual-encoder and cross-encoder, and explain the reasons for the formation of their similarity distribution characteristics in Sec \ref{sec:revisit}. Then, we identify the source of the performance gap between them, underscore the importance of hard negatives ranking, and elaborate on our proposed Contrastive Partial Ranking Distillation (CPRD) method in Sec \ref{sec:prd}. Finally, we present the overall training objective of dual-encoder in Sec \ref{sec:train}.

\vspace{-0.1cm}
\subsection{Dual-Encoder and Cross-Encoder}
\vspace{-0.1cm}
\label{sec:revisit}

\noindent \textbf{Dual-Encoder.} 
The dual-encoder takes an image and its paired text as input, and extracts global visual and textual representations $\bm{v}_i$ and $\bm{t}_i$ with separate encoders.
The dot product of $\bm{v}_i$ and $\bm{t}_i$ is used to measure the image-text similarity. 
During pre-training, following previous works\cite{li2021align}, we maintain two queues $\mathcal{Q}^v$ and $\mathcal{Q}^t$ to preserve the momentum features from current mini-batch $\{\bm{\hat{v}}_j\}_{j=1}^{B}$ and $\{\bm{\hat{t}}_j\}_{j=1}^{B}$ and previous iterations $\{\bm{\hat{v}}_j\}_{j=B+1}^{B + Nq}$ and $\{\bm{\hat{t}}_j\}_{j=B+1}^{B + Nq}$. 
For every image/text in the current mini-batch, its related text/image is deemed a positive sample, while the unmatched texts/images within the mini-batch and all samples in the previous iterations are considered as negatives. 
The contrastive learning (\textit{i.e.}, InfoNCE loss) is employed to maximize the similarity between positive image-text pairs while minimizing the similarity between negative pairs, which is formulated as:
\vspace{-0.3cm}
\begin{equation}
\vspace{-0.3cm}
\begin{aligned}
    \mathcal{L}_{\mathrm{I2T}}&=-\frac{1}{B}\sum_{i=1}^B \mathrm{log}\frac{\mathrm{exp}({\bm{v}_i^\top}\bm{\hat{t}}_i/ \tau)}{\sum_{j=1}^{B+N_q} \mathrm{exp}({\bm{v}_i^\top}\bm{\hat{t}}_j/ \tau)}, \\
    \mathcal{L}_{\mathrm{T2I}}&=-\frac{1}{B}\sum_{i=1}^B \mathrm{log}\frac{\mathrm{exp}({\bm{t}_i^\top}\bm{\hat{v}}_i/ \tau)}{\sum_{j=1}^{B+N_q} \mathrm{exp}(\bm{t}_i^\top\bm{\hat{v}}_j / \tau)}, \\
\end{aligned}
\end{equation}
where $\tau$ is the temperature and $B$ is the batch size.
The total loss for image-text contrastive learning is defined as:
\vspace{-0.3cm}
\begin{equation}
\vspace{-0.3cm}
    \mathcal{L}_{\mathrm{align}} = (\mathcal{L}_{\mathrm{I2T}}+\mathcal{L}_{\mathrm{T2I}}) / 2.
\end{equation}
The characteristics of independent encoding for each modality in the dual-encoder and the objective of contrastive learning lead to a moderate similarity distribution.

\noindent \textbf{Cross-Encoder.} 
Cross-encoders typically start by utilizing visual and textual encoders to extract detailed image and text representations $\bm{V}_i$ and $\bm{T}_j$ from the input image-text pair. 
Then the obtained primary representations are fed into a multi-modal encoder that employs self-attention or cross-attention to enable cross-modal interactions, thereby calculating the similarity score $\bm{p}_{i,j} \in [0,1]$ for the input image-text pair. 
The image-text matching task serves as the training objective for the cross-encoder, which aims at judging whether the input image-text pair is matched or not and is formulated as:
\vspace{-0.3cm}
\begin{equation}
\vspace{-0.3cm}
    \mathcal{L}_{itm} = - \frac{1}{|\mathcal{S}|}\sum_{(i,j)\in \mathcal{S}}y_{ij}log\bm{p}_{i,j} + (1 - y_{ij}) log(1-\bm{p}_{i,j}),\notag
\end{equation}
where $y_{ij} \in \{0, 1\}$ is the ground-truth label indicating whether the image-text pair is matched or not. $\mathcal{S}$ is the index set of image-text pairs, obtained by random sampling or hard negative sampling strategies.
The objective of image-text matching pushes $\bm{p}_{i,j}$ to either 0 or 1, making the similarity distribution of cross-encoder more concentrated. The substantial difference between similarity distribution of dual-encoder and cross-encoder makes KL-divergence based distillation less effective.

\vspace{-0.1cm}
\subsection{Contrastive Partial Ranking Distillation}
\vspace{-0.1cm}
\label{sec:prd}

To achieve effective knowledge transfer from cross-encoder to dual-encoder, it is essential to first identify the source of performance gap between dual-encoder and cross-encoder. 
We test the performance of the pre-trained dual-encoder and cross-encoder from \cite{li2022blip} on MSCOCO\cite{lin2014microsoft} dataset. Note that the retrieval results of cross-encoder are obtained by re-ranking top-K retrieved items from dual-encoder.
The results are presented in Table \ref{tab:comparison_DE_CE}, and we observe that (1) As K increases from 0 to 32, the retrieval performance significantly improves.
(2) When more less-challenging negatives are introduced, \ie, K continues to increase from 32 to 256, the performance saturates.
It indicates that dual-encoder has filtered easy negatives and ranked most relevant items (including positives and hard negatives) into top 32, but lacks accurate ranking of them. Furthermore, given the orginal loss of the dual-encoder (InfoNCE loss) encompasses the objective of ranking positives ahead of hard negatives, the distillation only needs to focus on the cross-encoder's ranking knowledge of hard negatives.

\begin{table}[t]
 \caption{Performance comparison between dual-encoder (DE) and cross-encoder (CE) with different candidate number $\mathrm{K}$.}
 \label{tab:comparison_DE_CE}
 \vspace{-0.25cm}
 \footnotesize
 \resizebox{\linewidth}{!}{
 \begin{tabular}{lccccccc}
    \toprule
    \multirow{2}{*}{Model} & \multicolumn{3}{c}{image$\rightarrow$text} & \multicolumn{3}{c}{text$\rightarrow$image} & \multirow{2}{*}{R@S}\\
     & R@1 & R@5 & R@10 & R@1 & R@5 & R@10 & \\
    \midrule
    DE & 73.8 & 92.1 & 95.7 & 55.8 & 80.6 & 88.1 & 486.2 \\
    \midrule
    CE, K=0 & 73.8 & 92.1 & 95.7 & 55.8 & 80.6 & 88.1 & 486.2 \\
    CE, K=4 & 76.6 & 91.3 & 95.6 & 60.6 & 80.5 & 87.8 & 492.4 \\
    CE, K=16 & 78.1 & 93.0 & 95.8 & 61.4 & 83.5 & 89.3 & 501.1 \\
    CE, K=32 & 78.4 & 93.4 & 96.4 & \textbf{61.5} & \textbf{83.8} & \textbf{90.0} & 503.4 \\
    CE, K=128 & 78.6 & 93.9 & 96.9 & 61.4 & 83.7 & 89.9 & \textbf{504.4} \\
    CE, K=256 & \textbf{78.5} & \textbf{93.8} & \textbf{96.9} & 61.4 & 83.7 & 89.9 & 504.2 \\    
 \bottomrule
\end{tabular}}
\vspace{-0.7cm}
\end{table}

Based on the above findings, we propose Contrastive Partial Ranking Distillation (CPRD), a method that leverages contrastive learning to enforce consistency in the ranking of hard negatives between dual-encoders and cross-encoders.
As shown in Figure \ref{fig:framework}, in order to transfer knowledge about hard negatives ranking effectively, our CPRD method first mining sufficient hard negative samples with dual-encoder's similarity score, then construct the ranking target of hard negatives with cross-encoder, and finally perform partial ranking distillation via contrastive learning to transfer cross-modal matching knowledge effectively.

\vspace{-1em}
\subsubsection{Hard Negative Mining} 
\vspace{-0.6em}

A straightforward approach to obtain hard negatives is selecting the most similar unmatched images/texts within the mini-batch.
However, the batch size must be sufficiently large to find adequately challenging hard negatives. 
Therefore, we expand the mining scope to the queues $\mathcal{Q}^v$ and $\mathcal{Q}^t$.
Here, we demonstrate the process of image-to-text retrieval, while the text-to-image retrieval is performed symmetrically.
For the $i$-th image, we use dual-encoder to compute its similarity scores with all negative texts in $\mathcal{Q}^t$, which contains samples from both current mini-batch and previous iterations. 
We then obtain the ranking result for the negative texts, denoted as $\bm{d}_i=\{d_{ij}\}_{j=1}^{B+N_q-1}$, by sorting the similarity scores in the descending order, where:
\vspace{-0.2cm}
\begin{equation}
\vspace{-0.2cm}
    \bm{v}_i^\top\hat{\bm{t}}_{d_{ij}} > \bm{v}_i^\top\hat{\bm{t}}_{d_{ik}},\ \forall \ j,k \in [1,...,B+N_q-1],\ j < k,
\end{equation}
and indices of top-$\mathrm{K}$ hard negative texts is $\bm{h}_i = \{d_{ij}\}_{j=1}^{\mathrm{K}}$.

\vspace{-0.2cm}
\subsubsection{Ranking Target Construction}\label{sec:ranking_target_construction}
\vspace{-0.1cm}

Then we construct the ranking target of top-K hard negatives with cross-encoder. To compute the similarity scores of the top-K hard negatives with the cross-encoder, we also need to maintain two additional queues $\mathcal{Q}_c^v$ and $\mathcal{Q}_c^t$ with size $N_c$ to preserve the image and text inputs from mini-batch and previous iterations for the cross-encoder.
The similarity score set $\bm{P}_{i}$ of the cross-encoder for all hard negative texts is denoted as $\bm{P}_{i} = \{\bm{p}_{i,d_{ij}} | j \in [1,2,...,\mathrm{K}]\}$.
The cross-encoder's ranking of hard negative texts $\bm{c}_i=\{c_{ij}\}_{j=1}^{\mathrm{K}}$ can then be obtained by sorting $\bm{P}_{i}$, where $c_{ij}$ satisfies:
\vspace{-0.2cm}
\begin{equation}
\vspace{-0.2cm}
    \bm{p}_{i,c_{ij}} > \bm{p}_{i,c_{ik}},\ \forall \ j,k \in [1,...,\mathrm{K}],\ j < k.
\end{equation}

However, the cross-encoder's ranking $\bm{c}_i$ can not be used as target directly because not all information in the ranking $\bm{c}_i$ is beneficial for dual-encoder. There are some low-ranked negative texts in $\bm{c}_i$ that have small similarity score $\bm{p}_{i,c_{ij}}$. We refer these texts as invalid hard negative texts $\{\bm{\hat{t}}_{c_{ij}}|\bm{p}_{i,c_{ij}} < m\}$, where $m$ is the hyper-parameter to distinguish between valid and invalid hard negative texts. The relative order among invalid hard negative texts does not contain helpful knowledge, and enforcing the dual-encoder to learn such relative order may hurt the model performance. Therefore, we construct a partial ranking $\bm{c}^*_i$ as the target, where the relative order between valid hard negative texts is the same with $\bm{c}_i$ but the relative order between invalid ones is disregarded.

It is worth noting that using the cross-encoder to calculate similarity scores online brings additional training costs. To address this issue, we can employ an offline approach. 
Specifically, we first use the dual-encoder to perform a fast retrieval on the entire pre-training dataset to obtain the top-N hard negative texts for each image. 
Then we compute similarity scores of these negative image-text pairs with cross-encoder and store them in a similarity bank.
During the training iterations, for the negative text $\bm{\hat{t}}_{d_{ij}}$, if $\bm{p}_{i,d_{ij}}$ is in the similarity bank, we load the pre-computed value; otherwise, we term $\bm{\hat{t}}_{d_{ij}}$ as invalid negative texts.
We set $\mathrm{N}\gg K$ to ensure that most valid hard negatives are considered.

\vspace{-0.2cm}
\subsubsection{Contrastive Partial Ranking Learning}
\vspace{-0.1cm}

Given the ranking target $\bm{c}^*_i$ from cross-encoder, we expect the rankings from dual-encoder ($\bm{h}_i$) can be \textbf{\textit{partial}} consistent with $\bm{c}^*_i$, which means only the relative order between valid hard negatives should be maintained and the relative order between invalid hard negatives is ignored. The learning objective is formulated as:
\vspace{-0.3cm}
\begin{equation}
\vspace{-0.3cm}
    \min\limits_{\theta} \mathrm{Dist}(\bm{h}_i, \bm{c}^*_i),
\end{equation}
where $\theta$ is the parameters of dual-encoder, $\mathrm{Dist}(\cdot, \cdot)$ is the metric to measure the partial consistency between two ranking results. 
Optimizing this objective faces two challenges: (1) An appropriate metric to measure the partial consistency between rankings; (2) The sort operation used to compute $\bm{c}^*_i$ and $\bm{h}_i$ is non-differentiable, hindering the end-to-end training. Both are hard to tackle.

To implement effective partial ranking learning, we transform the non-differentiable ranking mimicking into the optimization of relative similarity scores.
Specifically, for each valid hard negative text in $\bm{c}^*_i$, we require its similarity with the $i$-th image computed by dual-encoder to be larger than the similarities of other negative texts ranked behind it. 
For invalid hard negative texts, we do not impose this constraint, which avoids learning relative order between them.
The new objective can be formulated as:
\vspace{-0.3cm}
\begin{equation}
\vspace{-0.3cm}
    \begin{aligned}
    \label{equ:obj}
        \bm{v}^\top_i \bm{t}_{c_{ij}} &> \bm{v}^\top_i \bm{t}_{c_{ik}}, \\ 
        \forall j, k,\ \ j < J_i^*, &1 \leq j < k \leq \mathrm{K}, \\ 
    \end{aligned}
\end{equation}
where $J_i^*$ denotes the index of the first invalid hard negative text in the $\bm{c}^*_{i}$.

Although Equation \ref{equ:obj} can already be used as a differentiable loss function for optimizing the dual-encoder, we find that an inappropriate loss function type may interfere with the coarse image-text alignment learned by dual-encoder. Formulating the ranking learning process in the form of contrastive learning is more coordinated with training of dual-encoder and yields better results. 
Therefore, we propose a new contrastive partial ranking loss based on the InfoNCE loss, which is formulated as:
\vspace{-0.3cm}
\begin{flalign}
\vspace{-0.7cm}
    \mathcal{L}_{ij} = -log \frac{\mathrm{exp}(\bm{v}^\top_i\bm{\hat{t}}_{c_{ij}} / \tau)}{\sum \limits_{k=j}^{\mathrm{K}} \mathrm{exp}(\bm{v}^\top_i\bm{\hat{t}}_{c_{ik}} / \tau) + \sum \limits_{k=\mathrm{K}+1}^{B+N_q-1} \mathrm{exp}(\bm{v}^\top_i\bm{\hat{t}}_{d_{ik}} / \tau)}, \notag
\end{flalign}

\vspace{-0.5cm}
\begin{flalign}
\vspace{-0.4cm}
\mathcal{L}_{\mathrm{CPRD}}^{\mathrm{I2T}} = \frac{1}{B}\sum_{i=1}^{B} \frac{1}{J_i^*-1} \sum \limits_{j=1}^{J_i^*-1} \mathcal{L}_{ij}, &&
\end{flalign}
where $\mathcal{L}_{ij}$ has the same effect with Equation \ref{equ:obj}. As mentioned above, $\mathcal{L}_{\mathrm{CPRD}}^{\mathrm{T2I}}$ can be computed in the symmetrical way and the final contrastive partial ranking distillation loss $\mathcal{L}_{\mathrm{CPRD}}$ is defined as:
\vspace{-0.1cm}
\begin{equation}
\vspace{-0.1cm}
    \mathcal{L}_{\mathrm{CPRD}} = (\mathcal{L}_{\mathrm{CPRD}}^{\mathrm{I2T}} + \mathcal{L}_{\mathrm{CPRD}}^{\mathrm{T2I}}) / 2.
\end{equation}

\subsection{Training Objectives}
\label{sec:train}
We perform distillation training with both image-text contrastive loss and new contrastive partial ranking distillation loss to learn the coarse image-text alignment and detailed ranking between hard negatives simultaneously. 
The total loss is formulated as:
\vspace{-0.1cm}
\begin{equation}
\vspace{-0.1cm}
    \mathcal{L} = \mathcal{L}_{\mathrm{align}} + \mathcal{L}_{\mathrm{CPRD}}.
\end{equation}

\section{Experiments}
\subsection{Datasets}

\noindent \textbf{Pre-training Datasets.} 
We pre-train our model with CC4M dataset, which contains 4 million images and 5.1 million captions from four public image-text datasets, including Conceptual Captions 3M\cite{sharma2018conceptual}, SBU\cite{ordonez2011im2text}, MSCOCO\cite{lin2014microsoft} and Visual Genome\cite{krishna2017visual}.

\noindent \textbf{Downstream Datasets.}
We conduct downstream image-text retrieval evaluation on two widely used datasets: MSCOCO\cite{lin2014microsoft} and Flick30K\cite{plummer2015flickr30k}. 
In addition, we validate the effectiveness of our method on improving the ranking ability with the CrissCrossed Caption\cite{parekh2021crisscrossed} dataset. 
The details of these downstream datasets and the evaluation metrics can be found in the supplemental material.

\subsection{Implementation Details.}
For dual-encoder, we adopt $\mathrm{BERT_{base}}$\cite{kenton2019bert} as text encoder and a ViT-B/16\cite{dosovitskiy2020image} pre-trained on ImageNet-1k as the visual encoder. For cross-encoder, we experiment with three models with varying performance, including ALBEF\cite{li2021align} pre-trained on 5M image-text pairs, ALBEF pre-trained on 15M image-text pairs, and BLIP\cite{li2022blip} pre-trained on $>$200M image-text pairs. We use the AdamW\cite{loshchilov2018decoupled} optimizer with a weight decay of 0.02. The learning rate is warmed up to 3$e{-4}$ in the first 2000 iterations and decays to 1$e{-5}$ following a cosine schedule. We pre-train the model for 20 epochs with a batch size of 512 on 8 NVIDIA V100 GPUs. 
We take the image resolution of 256$\times$256 for pre-training and increase the image resolution to 384$\times$384 for fine-tuning.
The momentum coefficient for updating momentum encoders is set as 0.995. The number of hard negative K and the threshold $m$ is set as 16 and 0.75. The queue size $N_q$ and $N_c$ are set as 57856 and 16384 respectively.
The learnable temperature hyper-parameter for contrastive loss is initialized to 0.07. More implementation details can be found in the supplementary materials.

\subsection{Teach Dual-Encoder with Cross-Encoder}

In this section, we provide detailed steps to demonstrate how to use the cross-encoder as teacher for the dual-encoder effectively. 
All models in this section are pre-trained on CC3M using an image resolution of $224\times 224$ with 8 NVIDIA V100 GPUs, and then tested for zero-shot image-text retrieval performance on MSCOCO. 
The teacher model is ALBEF pre-trained on CC3M and is kept frozen during the distillation process, while the student model is a dual-encoder trained from scratch.

\begin{table}[t]
 \caption{The performance comparison with different queue size $N_c$. ``DE" and ``CE" represent the baseline dual-encoder and teacher cross-encoder.}
 \label{tab:comparison_queue_size}
 \vspace{-0.4cm}
 \footnotesize
 \resizebox{\linewidth}{!}{
 \begin{tabular}{lccccccc}
    \toprule
    \multirow{2}{*}{$N_c$} & \multicolumn{3}{c}{image$\rightarrow$text} & \multicolumn{3}{c}{text$\rightarrow$image} & \multirow{2}{*}{R@S}\\
     & R@1 & R@5 & R@10 & R@1 & R@5 & R@10 & \\
    \midrule
    DE & 32.0 & 59.4 & 71.5 & 24.4 & 49.5 & 61.0 & 297.8 \\
    CE & 40.5 & 66.5 & 76.1 & 30.2 & 55.3 & 66.5 & 335.1 \\
    \hline
    0 & 32.6 & 60.1 & 71.0 & 24.7 & 49.8 & 61.7 & 299.9 \\
    1024 & 32.6 & 60.3 & 71.9 & 25.9 & 51.0 & 62.7 & 304.4 \\
    4096 & 33.7 & 60.6 & 71.7 & 26.9 & 52.3 & \underline{64.0} & 309.2 \\
    16384 & \underline{34.3} & \underline{61.4} & \underline{73.2} & \underline{27.0} & \textbf{52.8} & \textbf{64.5} & \underline{313.2} \\
    32768 & \textbf{34.6} & \textbf{62.6}  & \textbf{73.5}  & \textbf{27.3}  & \underline{52.4}  & 63.8 & \textbf{314.2} \\
 \bottomrule
\end{tabular}}
\vspace{-0.4cm}
\end{table}

\begin{table}[t]
 \caption{The performance comparison with different numbers of hard negative considered for contrastive partial ranking distillation. ``None" indicates the offline approach for calculating hard negative pairs' similarity with cross-encoder, which does not need to set K.}
 \label{tab:comparison_K}
 \vspace{-0.4cm}
 \footnotesize
 \resizebox{\linewidth}{!}{
 \begin{tabular}{lccccccc}
    \toprule
    \multirow{2}{*}{K} & \multicolumn{3}{c}{image$\rightarrow$text} & \multicolumn{3}{c}{text$\rightarrow$image} & \multirow{2}{*}{R@S}\\
     & R@1 & R@5 & R@10 & R@1 & R@5 & R@10 & \\
    \midrule
    0 & 32.0 & 59.4 & 71.5 & 24.4 & 49.5 & 61.0 & 297.8 \\
    2 & 32.4 & 60.7 & 71.8 & 24.6 & 49.6 & 61.4 & 300.5 \\
    4 & 33.5 & 61.1 & \underline{72.9} & 25.9 & 50.9 & 62.2 & 306.5 \\
    8 & \textbf{34.6} & \underline{61.3} & \textbf{73.2} & 26.7 & 51.6 & 63.5 & 310.9 \\
    16 & 34.1 & 61.1 & 72.6 & \textbf{27.3} & \textbf{53.0} & \underline{64.3} & \underline{312.4} \\
    32 & \underline{34.3} & \underline{61.4} & \textbf{73.2} & \underline{27.0} & \underline{52.8} & \textbf{64.5} & \textbf{313.2} \\
    None & 34.0 & 60.8 & 72.5 & 27.3 & 52.7 & 64.0 & 311.3 \\
 \bottomrule
\end{tabular}}
\vspace{-0.45cm}
\end{table}

\noindent \textbf{The effect of queue size $N_c$.} 
The queue size $N_c$ determines the number of negative samples, which affects the number of valid hard negatives and the final distillation performance. Therefore, choosing a sufficiently large queue size is crucial. As shown in Table \ref{tab:comparison_queue_size}, when we set the queue size to 0, only the negative samples in the current batch are utilized for distillation, resulting in a limited number of valid hard negatives and less valuable knowledge. Consequently, the performance improvement compared to the baseline is relatively weak. When we increase the queue size to 1024, the number of valid hard negatives increases, and the model performance achieve significant improvement. As the queue size further increases, the model performance gradually improves and tends to saturate. This is because, in the training dataset, the number of valid hard negatives for a specific sample is limited. Further increasing the size of the queue leads to a saturation in the number of valid hard negatives. We set the queue size to 16384 by default.

\begin{table}[t]
 \caption{The performance comparison with different valid hard negative threshold $m$.}
 \label{tab:comparison_threshold}
 \vspace{-0.35cm}
 \footnotesize
 \resizebox{\linewidth}{!}{
 \begin{tabular}{lccccccc}
    \toprule
    \multirow{2}{*}{$m$} & \multicolumn{3}{c}{image$\rightarrow$text} & \multicolumn{3}{c}{text$\rightarrow$image} & \multirow{2}{*}{R@S}\\
     & R@1 & R@5 & R@10 & R@1 & R@5 & R@10 & \\
    \midrule
    1.0 & 32.0 & 59.4 & 71.5 & 24.4 & 49.5 & 61.0 & 297.8 \\
    0.9 & 33.3 & 60.7 & 72.0 & \underline{26.8} & \underline{52.6} & \underline{64.2} & 309.6\\
    0.75 & \textbf{34.3} & \underline{61.4} & \textbf{73.2} & \textbf{27.0} & \textbf{52.8} & \textbf{64.5} & \textbf{313.2} \\
    0.5 & \underline{33.4} & \textbf{61.5} & \underline{73.0} & \textbf{27.0} & 52.3 & 63.6 & \underline{310.8} \\
    0.0 & 30.8 & 57.2 & 69.4 & 24.1 & 48.4 & 60.5 & 290.4 \\
 \bottomrule
\end{tabular}}
\vspace{-0.4cm}
\end{table}

\begin{table}[t]
 \caption{The performance comparison with different distillation methods.}
 \label{tab:distill_comp}
 \vspace{-0.35cm}
 \footnotesize
 \resizebox{\linewidth}{!}{
 \begin{tabular}{lccccccc}
    \toprule
    \multirow{2}{*}{Method} & \multicolumn{3}{c}{image$\rightarrow$text} & \multicolumn{3}{c}{text$\rightarrow$image} & \multirow{2}{*}{R@S}\\
     & R@1 & R@5 & R@10 & R@1 & R@5 & R@10 & \\
    \midrule
    None & 32.0 & 59.4 & 71.5 & 24.4 & 49.5 & 61.0 & 297.8 \\
    KL & 32.5 & 59.6 & 71.4 & 26.0 & 50.9 & 62.9 & 303.3 \\
    M3SE & 30.8 & 57.2 & 69.4 & 24.1 & 48.4 & 60.5 & 290.4 \\
    R-M3SE & 32.8 & 60.4 & 71.0 & 25.6 & 50.5 & 62.0 & 302.3 \\
    CPRD$_{m^*}$ & \underline{33.4} & \underline{60.8} & \underline{72.4} & \underline{26.3} & \underline{51.5} & \underline{63.4} & \underline{307.7} \\
    CPRD & \textbf{34.3} & \textbf{61.4} & \textbf{73.2} & \textbf{27.0} & \textbf{52.8} & \textbf{64.5} & \textbf{313.2} \\
 \bottomrule
\end{tabular}}
\vspace{-0.7cm}
\end{table}

\begin{table*}[t]
 \caption{Comparative results for fine-tuned image-text retrieval results on the Flickr30K (1K) test set and MSCOCO (5K) test set. We make comparisons with dual-encoder methods and cross-encoder methods. ``DE" and ``CE" represent the baseline dual-encoder and the teacher cross-encoder. Our method improves the performance of baseline model significantly, surpasses previous state-of-the-art dual-encoder methods by a large margin, and achieves comparable performance with some cross-encoder methods while keeping the high retrieval efficiency. \textbf{Higher} R@K indicates better performance. \textbf{PT Pairs}: the number of image-text pairs for pre-training. $\mathbf{\dag}$ is ensemble result of two models. $*$ models use 940M tagged images for visual encoder pre-training.}
 \vspace{-0.3cm}
 \label{tab:comparison_sota}
 \resizebox{\textwidth}{!}{
 \begin{tabular}{lccccccccccccccc}
    \toprule
    \multirow{3}{*}{Model} & \multirow{3}{*}{PT Pairs} & \multicolumn{7}{c}{Flickr30K (1K test set)} & \multicolumn{7}{c}{MSCOCO (5K test set)} \\
    & & \multicolumn{3}{c}{image$\rightarrow$text} & \multicolumn{3}{c}{text$\rightarrow$image} & \multirow{2}{*}{R@S} & \multicolumn{3}{c}{image$\rightarrow$text} & \multicolumn{3}{c}{text$\rightarrow$image} & \multirow{2}{*}{R@S} \\
    & & R@1 & R@5 & R@10 & R@1 & R@5 & R@10 &  & R@1 & R@5 & R@10 & R@1 & R@5 & R@10 &  \\
    \midrule
    \multicolumn{16}{l}{\textbf{Cross-Encoder}} \\
    Pixel-BERT-X152 \cite{huang2020pixel} & 5.6M & 87.0 & 98.9 & 99.5 & 71.5 & 92.1& 95.8 & 544.8 & 63.6 & 87.5 & 93.6 & 50.1 & 77.6 & 86.2 & 458.6\\
    VILLA-base \cite{gan2020large} & 9.6M & 86.6 & 97.9 & 99.2 & 74.7 & 92.9 & 95.8 & 547.1& -- & -- & -- & -- & -- & -- & --\\
    Oscar-base \cite{li2020oscar} & 6.5M & -- & -- & -- & -- & -- & -- & -- & 70.0 & 91.1 & 95.5 & 54.0 & 80.8 & 88.5 & 479.9\\
    ViLT \cite{kim2021vilt} & 9.9M & 83.5 & 96.7 & 98.6 & 64.4 & 88.7 & 93.8 & 525.7 & 61.5 & 86.3 & 92.7 & 42.7 & 72.9 & 83.1 & 439.2\\
    VinVL-base \cite{zhang2021vinvl} & 8.9M & -- & -- & -- & -- & -- & -- & -- & \textbf{74.6} & \textbf{92.6} & \textbf{96.3} & \textbf{58.1} & \textbf{83.2} & \textbf{90.1} & \textbf{494.9}\\
    ALBEF \cite{li2021align} & 5.1M & \textbf{94.3} & \textbf{99.4} & \textbf{99.8} & \textbf{82.8} & \textbf{96.7} & \textbf{98.4} & \textbf{571.4} & 73.1 & 91.4 & 96.0 & 56.8 & 81.5 & 89.2 & 488.0\\
    \midrule
    \multicolumn{16}{l}{\textbf{Dual-Encoder}} \\
    \textcolor{grayfont}{VSE$\infty^{*\dag}$} \textcolor{grayfont}{\cite{chen2021learning}} & \textcolor{grayfont}{--} & \textcolor{grayfont}{88.7} & \textcolor{grayfont}{98.9} & \textcolor{grayfont}{99.8} & \textcolor{grayfont}{76.1} & \textcolor{grayfont}{94.5} & \textcolor{grayfont}{97.1} & \textcolor{grayfont}{555.1} & \textcolor{grayfont}{68.1} & \textcolor{grayfont}{90.2} & \textcolor{grayfont}{95.2} & \textcolor{grayfont}{52.7} & \textcolor{grayfont}{80.2} & \textcolor{grayfont}{88.3} &  \textcolor{grayfont}{474.7}\\
    \textcolor{grayfont}{COOKIE$^{*\dag}$ \cite{wen2021cookie}} & \textcolor{grayfont}{5.9M} & \textcolor{grayfont}{89.0} & \textcolor{grayfont}{98.9} & \textcolor{grayfont}{99.7} & \textcolor{grayfont}{75.6} & \textcolor{grayfont}{94.6} & \textcolor{grayfont}{97.2} & \textcolor{grayfont}{555.0} & \textcolor{grayfont}{71.6} & \textcolor{grayfont}{90.9} &  \textcolor{grayfont}{95.4} & \textcolor{grayfont}{54.5} & \textcolor{grayfont}{81.0} & \textcolor{grayfont}{88.2} & \textcolor{grayfont}{481.6}\\
    LightningDOT \cite{sun2021lightningdot} & 9.5M & 83.9 & 97.2 & 98.6 & 69.9 & 91.1 & 95.2 & 535.9 & 60.1 & 85.1 & 91.8 & 45.8 & 74.6 & 83.8 & 441.2\\
    COOKIE \cite{wen2021cookie} & 5.9M & 84.7 & 96.9 & 98.3 & 68.3 & 91.1 & 95.2 & 534.5 & 61.7 & 86.7 & 92.3 & 46.6 & 75.2 & 84.1 & 446.6\\
    COTS \cite{lu2022cots} & 5.3M & 88.2 & 98.5 & 99.7 & 75.2 & 93.6 & 96.5 & 551.7 & 66.9 & 88.8 & 94.0 & 50.5 & 77.6 & 86.1 & 463.9\\
    COTS \cite{lu2022cots} & 15.3M & 90.6 & 98.7 & 99.7 & 76.5 & 93.9 & 96.6 & 556.0 & 69.0 & 90.4 & 94.9 & 52.4 & 79.0 & 86.9 & 472.6\\
    \hline
    \rowcolor{mygray}
    DE & 5.1M & 89.1 & 98.8 & 99.7 & 74.1 & 92.8 & 96.2 & 550.7 & 66.7 & 88.9 & 94.4 & 48.5 & 76.7 & 85.1 & 460.3 \\
    \rowcolor{mygray}
    +CPRD & 5.1M & \textbf{90.8} & 98.7 & 99.4 & 76.4 & 93.9 & 96.8 & 556.0 & 69.7 & 91.1 & 95.8 & 52.1 & 78.9 & 86.8 & 474.4\\
    \rowcolor{mygray}
    \textcolor{grayfont}{CE} & \textcolor{grayfont}{5.1M} & \textcolor{grayfont}{94.5} & \textcolor{grayfont}{99.0} & \textcolor{grayfont}{99.7} & \textcolor{grayfont}{78.6} & \textcolor{grayfont}{94.3} & \textcolor{grayfont}{97.1} & \textcolor{grayfont}{563.2} & \textcolor{grayfont}{74.4} & \textcolor{grayfont}{92.3} & \textcolor{grayfont}{96.5} & \textcolor{grayfont}{58.0} & \textcolor{grayfont}{82.8} & \textcolor{grayfont}{89.7} & \textcolor{grayfont}{493.7} \\
    \hline
    \rowcolor{mygray}
    +CPRD & 5.1M & 90.7 & \textbf{99.0} & \textbf{99.7} & \textbf{78.6} & \textbf{94.9 }& \textbf{97.4} & \textbf{560.3} & \textbf{70.8} & \textbf{91.7} & \textbf{96.2} & \textbf{53.4} & \textbf{80.6} & \textbf{88.5} & \textbf{481.2}\\
    \rowcolor{mygray}
    \textcolor{grayfont}{CE} & \textcolor{grayfont}{15.2M} & \textcolor{grayfont}{95.9} & \textcolor{grayfont}{99.8} & \textcolor{grayfont}{100.} & \textcolor{grayfont}{85.6} & \textcolor{grayfont}{97.5} & \textcolor{grayfont}{98.9} & \textcolor{grayfont}{577.7} & \textcolor{grayfont}{77.6} & \textcolor{grayfont}{94.3} & \textcolor{grayfont}{97.2} & \textcolor{grayfont}{60.7} & \textcolor{grayfont}{84.3} & \textcolor{grayfont}{90.5} & \textcolor{grayfont}{504.6} \\
 \bottomrule
\end{tabular}}
\vspace{-0.6cm}
\end{table*}

\noindent \textbf{The effect of hard negatives number K.} 
Similarly, the choice of K also affects the number of valid hard negatives and the effectiveness of knowledge distillation. 
As shown in Table \ref{tab:comparison_K}, when set K to 0, we do not consider the ranking of any negative samples, causing the model to degrade to the baseline model. Gradually increasing K leads to more valid hard negative samples being considered, making the distillation more effective and the model performance improves. When K is increased to 32, the model performance tends to saturate, as most of the valid hard negatives are considered. Moreover, increasing K results in heavier computation, thus we set K to 16 by default. It is worth noting that we can calculate the similarity of hard negative pairs in an offline manner to avoid extra training costs. For each image/text in the training dataset, we use the cross-encoder to calculate the scores of the top-1000 hard negative sample pairs as mentioned in Section \ref{sec:ranking_target_construction}. As shown in Table \ref{tab:comparison_K}, the offline calculation approach achieves comparable performance with the online approach. It should be noted that when using the offline approach, we directly search for valid hard negatives from the current batch and queue without the need of setting K.

\noindent \textbf{The effect of valid hard negative threshold \textit{m}.} The threshold $m$ controls the difficulty of negative samples for which dual-encoder needs to learn their relative order. As shown in Table \ref{tab:comparison_threshold}, when setting $m=0$, all relative order among top-$\mathrm{K}$ hard negatives are learned by dual-encoder, and the performance declines significantly because dual-encoder is also enforced to learn the relative order between easy negatives, which contains no valid knowledge and introduces interference for the dual-encoder. When setting $m=1$, none of negative samples are considered in the partial ranking learning, and thus the model degenerates into the baseline model. We also empirically test with $m=0.5, 0.75, 0.9$. We find that the performance of all three models is better than the baseline model, showing the robustness of our method toward the threshold $m$. The best result is achieved when $m=0.75$. Therefore, we set the default value of $m$ to $0.75$. These experimental results validate our findings that only relative order between valid hard negatives conveys valuable knowledge.

\begin{table}[t]
 \caption{Comparison for image-text retrieval results (without fine-tuning) on the MSCOCO (5K) test set.}
 \vspace{-0.3cm}
 \label{tab:comparison_wo_ft_coco}
 \resizebox{\linewidth}{!}{
 \begin{tabular}{lccccccc}
    \toprule
    \multirow{2}{*}{Model} & \multicolumn{3}{c}{image$\rightarrow$text} & \multicolumn{3}{c}{text$\rightarrow$image} & \multirow{2}{*}{R@S}\\
     & R@1 & R@5 & R@10 & R@1 & R@5 & R@10 & \\
    \midrule
    \multicolumn{8}{l}{\textbf{Pre-train from Scratch}} \\
    CLIP \cite{radford2021learning} & 58.4 & 81.5 & 88.1 & 37.8 & 62.4 & 72.2 & 400.4 \\
    ALIGN \cite{jia2021scaling} & 58.6 & 83.0 & 87.9 & \textbf{45.6} & 69.8 & 78.6 & 423.5\\
    COTS \cite{lu2022cots} & 60.4 & 84.7 & 91.7 & 43.8 & \underline{71.6} & \underline{81.3} & 433.5 \\
    
    \rowcolor{mygray}
    DE & \underline{61.7} & \underline{86.0} & \underline{92.3} & 41.5 & 70.4 & 80.6 & 432.5 \\
    \rowcolor{mygray}
    +CPRD & \textbf{62.4} & \textbf{86.5} & \textbf{93.0} & \underline{45.1} & \textbf{74.1} & \textbf{83.6} & \textbf{444.7} \\
    \rowcolor{mygray}
    \textcolor{grayfont}{CE} & \textcolor{grayfont}{73.2} & \textcolor{grayfont}{91.5} & \textcolor{grayfont}{95.7} & \textcolor{grayfont}{54.8} & \textcolor{grayfont}{79.7} & \textcolor{grayfont}{87.0} & \textcolor{grayfont}{481.9} \\
    \hline
    \multicolumn{8}{l}{\textbf{Post Pre-train}} \\
    \rowcolor{mygray}
    DE & 65.2 & 88.4 & 93.5 & 50.0 & 76.2 & 84.8 & 458.1 \\
    \rowcolor{mygray}
    +CPRD & \textbf{65.5} & \textbf{88.5} & \textbf{93.6} & \textbf{51.7} & \textbf{77.8} & \textbf{86.0} & \textbf{463.1} \\
    \rowcolor{mygray}
    \textcolor{grayfont}{CE} & \textcolor{grayfont}{71.9} & \textcolor{grayfont}{91.1} & \textcolor{grayfont}{95.2} & \textcolor{grayfont}{58.7} & \textcolor{grayfont}{81.7} & \textcolor{grayfont}{88.5} & \textcolor{grayfont}{487.1} \\
 \bottomrule
\end{tabular}}
\vspace{-0.7cm}
\end{table}

\noindent \textbf{Comparison with Different Distillation Methods.} Based on the optimal setting identified through the above experiments, we explore different distillation methods to transfer knowledge from to dual-encoder. KL is the KL-divergence loss which constrains the consistency between similarity score distribution. M3SE\cite{menon2022defense} requires that (1) the similarity margin between positive and hardest negative should be same for dual-encoder and cross-encoder, (2) similarity of other negative samples should be smaller than that of hardest negative. 

As shown in Table \ref{tab:distill_comp}, KL method achieves limited performance improvement due to the difference between similarity score distribution of dual-encoder and cross-encoder. M3SE even results in performance decline because the similarity range is different for dual-encoder and cross-encoder. To tackle this problem, we propose a modification R-M3SE to first re-scale similarity score with min-max normalization, and we observe the retrieval performance is improved but still worse than our method. To validate whether the form of loss function affects the distillation, we propose a variant of our method CPRD$_{m^*}$, which adjusts the threshold $m$ adaptively to ensure there always has only one valid hard negative. CPRD$_{m^*}$ has the highly-similar objective with R-M3SE but is implemented with infoNCE loss. It can be seen that CPRD$_{m^*}$ surpasses R-M3SE on all metrics, validating that it is beneficial to make the distillation loss coordinated with the training of dual-encoder.

\subsection{Comparison with SOTA}

\noindent \textbf{Image-Text Retrieval.} We compare with state-of-the-art image-text retrieval methods on Flickr30K and MSCOCO datasets. The experimental results under fine-tuning setting are shown in Table \ref{tab:comparison_sota}. Compared with our dual-encoder baseline, our proposed distillation method achieves significant improvement. On the Flickr30K dataset, we achieve higher performance by 1.7\% and 2.3\% on the R@1 of image-to-text and text-to-image retrieval. On the MSCOCO dataset, we also surpass the baseline by 3.0\% and 3.6\% on the R@1 of two retrieval tasks. We also experiment with a stronger cross-encoder trained with 15.2M data as teacher, the retrieval performance is further improved, showing that a stronger teacher can transfer more valid knowledge to student even with the same training data.

Under a fair comparison experimental setting (excluding VSE$\infty^{*\dag}$ and COOKIE$^{*\dag}$ as they use 940M tagged images for visual-encoder pre-training), our method also outperforms other dual-encoder methods by a large margin under all evaluation metrics. Specifically, compared with the current state-of-the-art dual-encoder method COTS \cite{lu2022cots} with 5.3M pre-training data, our method with similar data size achieves higher performance by 1.8\% and 1.6\% on the R@1 of MSCOCO dataset. Moreover, our method with 5.1M pre-training data outperforms COTS pre-trained on 15.3M image-text pairs. 
Furthermore, our method also achieves comparable performance with the cross-encoder methods VinVL-base and ALBEF while much more efficient. 

The experimental results without fine-tuning are shown in Table \ref{tab:comparison_wo_ft_coco}. Our distillation method improves the performance of dual-encoder on all evaluation metrics. With a similar pre-training data size, our method outperforms the COTS \cite{lu2022cots} by 2.0\% and 1.3\% on the R@1 of two retrieval tasks. Our method also outperforms CLIP \cite{radford2021learning} and achieves comparable performance with ALIGN \cite{jia2021scaling}, which utilize 78$\times$ and 356$\times$ pre-training data than our method respectively. Moreover, we also evaluate our method under post pre-train setting, where we initialize our dual-encoder with the pre-trained weights from BLIP-129M and continue to pre-train the model with image-text contrastive loss and our distillation loss. Our method can further improve the performance of the pre-trained model with only a small additional computation cost (10k iterations).

\begin{table}[t]
 \caption{Spearman's R Bootstrap Correlation ($\times$100) on Crisscrossed Captions dataset. ``FT" indicates the model is fine-tuned on the MSCOCO dataset. \textbf{STS}, \textbf{SIS} and \textbf{SITS} represent the task of semantic text similarity, semantic image similarity and semantic image-text similarity.}
 \vspace{-0.3cm}
 \label{tab:image_text_ranking}
 \resizebox{\linewidth}{!}{
 \begin{tabular}{lcccc}
    \toprule
    \multirow{2}{*}{Model} & \textbf{STS} & \textbf{SIS} & \textbf{SITS} & \multirow{2}{*}{\textbf{Mean Avg}}\\
     & avg $\pm$ std & avg $\pm$ std & avg $\pm$ std & \\
    \midrule
    VSE++\cite{faghrivse++} & \underline{74.4$\pm$0.4} & 73.3$\pm$0.9 & 55.2$\pm$1.5 & 67.6 \\
    VSRN \cite{li2019visual} & 73.0$\pm$0.4 & 70.1$\pm$1.0 & 60.4$\pm$1.3 & 67.8\\
    DE$\mathrm{_{I2T}}$ \cite{parekh2021crisscrossed} & 50.9$\pm$0.6 & \textbf{81.3$\pm$0.7} & 61.6$\pm$1.4 & 64.6 \\
    DE$\mathrm{_{T2T+I2T}}$\cite{parekh2021crisscrossed} & 74.2$\pm$0.4 & 74.5$\pm$0.9 & 61.9$\pm$1.3 & 70.2 \\
    ALIGN\cite{li2021align} & 72.9$\pm$0.4 & \underline{77.2$\pm$0.8} & \textbf{67.6$\pm$1.2} & \textbf{72.6} \\
    \hline
    \rowcolor{mygray}
    DE & 74.1$\pm$0.4 & 75.1$\pm$0.8 & 61.8$\pm$1.4 & 70.3 \\
    \rowcolor{mygray}
    +CPRD & \textbf{74.9$\pm$0.3} & 75.6$\pm$0.8 & 64.3$\pm$1.3 & 71.6 \\
    \rowcolor{mygray}
    \textcolor{grayfont}{CE} & \textcolor{grayfont}{--} & \textcolor{grayfont}{--} & \textcolor{grayfont}{67.3$\pm$1.2} & \textcolor{grayfont}{--} \\
    \hline
    \rowcolor{mygray}
    DE-FT & 73.9$\pm$0.4 & 75.7$\pm$0.8 & 65.1$\pm$1.3 & 71.5 \\
    \rowcolor{mygray}
    +CPRD-FT & \underline{74.4$\pm$0.4} & 76.1$\pm$0.9 & \underline{66.9$\pm$1.2} & \underline{72.5} \\
    \rowcolor{mygray}
    \textcolor{grayfont}{CE-FT} & \textcolor{grayfont}{--} & \textcolor{grayfont}{--} & \textcolor{grayfont}{69.4$\pm$1.1} & \textcolor{grayfont}{--} \\
 \bottomrule
\end{tabular}}
\vspace{-0.7cm}
\end{table}

\noindent \textbf{Image-Text Ranking.} To further validate the effectiveness of our CPRD method on improving the ranking ability of dual-encoder, we perform evaluation on the CrissCrossed caption dataset for image-text ranking task. We report the Spearman's R bootstrapped correlation for image-text ranking task, \ie, ``\textbf{SITS}" in Table \ref{tab:image_text_ranking}, which reflects the consistency between human ranking and model ranking. Compared with vanilla dual-encoder, our CPRD method achieves 2.5 improvement. The performance of our method is further improved after fine-tuning on the MSCOCO dataset, and achieves comparable results with ALIGN. In addition, we also report the results on text-text similarity ranking (\textbf{STS}) and image-image similarity ranking (\textbf{SIS}). It is worth noting that although our CPRD method only distills the knowledge of cross-modal matching, it can also improve the accuracy of measuring intra-modal similarity.

\section{Conclusion}
In this work, we investigate how to effectively distill cross-modal knowledge from cross-encoder to dual-encoder for image-text retrieval. 
We identify three key factors and propose a novel Contrastive Partial Ranking Distillation(CPRD) method. Our method focuses on learning relative order among valid hard negatives while disregarding relative order among invalid hard negatives and easy negatives. We implement our method with contrastive learning, which aligns with training of dual-encoder and transfers knowledge effectively without disrupting the learning of image-text alignment.
Comprehensive experiments on image-text retrieval and ranking show the superiority of our method compared to other distillation methods. Moreover, our method significantly improves retrieval and ranking accuracy of dual-encoder under various experiment settings.

\noindent \textbf{Acknowledgments}
\small{This work is supported by the National Science and Technology Major Project (Grant No.2022ZD0118501), Beijing Natural Science Foundation (Grant No. JQ21017, L223003), the Natural Science Foundation of China (Grant No. 62222206, 62036011, 62192782, 62302501, 62202470, U2033210), The Project of Beijing Science and technology Committee(Project No.Z231100005923046), Research Fund of ARC Lab, Tencent PCG.}

{
    \small
    \bibliographystyle{ieeenat_fullname}
    \bibliography{main}
}

\clearpage

\appendix

\renewcommand{\thesection}{\Alph{section}}

In this supplementary materials, we further explain the differences and connections between score distribution distillation and ranking distillation, in order to analyze the advantages of ranking distillation in the process of distilling knowledge from cross-encoder to dual-encoder. We also elaborate on (1) details about pre-training datasets, downstream datasets, and evaluation metrics of downstream tasks; (2) Visualizations about image-to-text retrieval and text-to-image retrieval. (3) More ablation study for CPRD loss.

\section{Score Distribution Distillation and Ranking Distillation}
\begin{figure}[h]
    \centering
    \includegraphics[width=\linewidth]{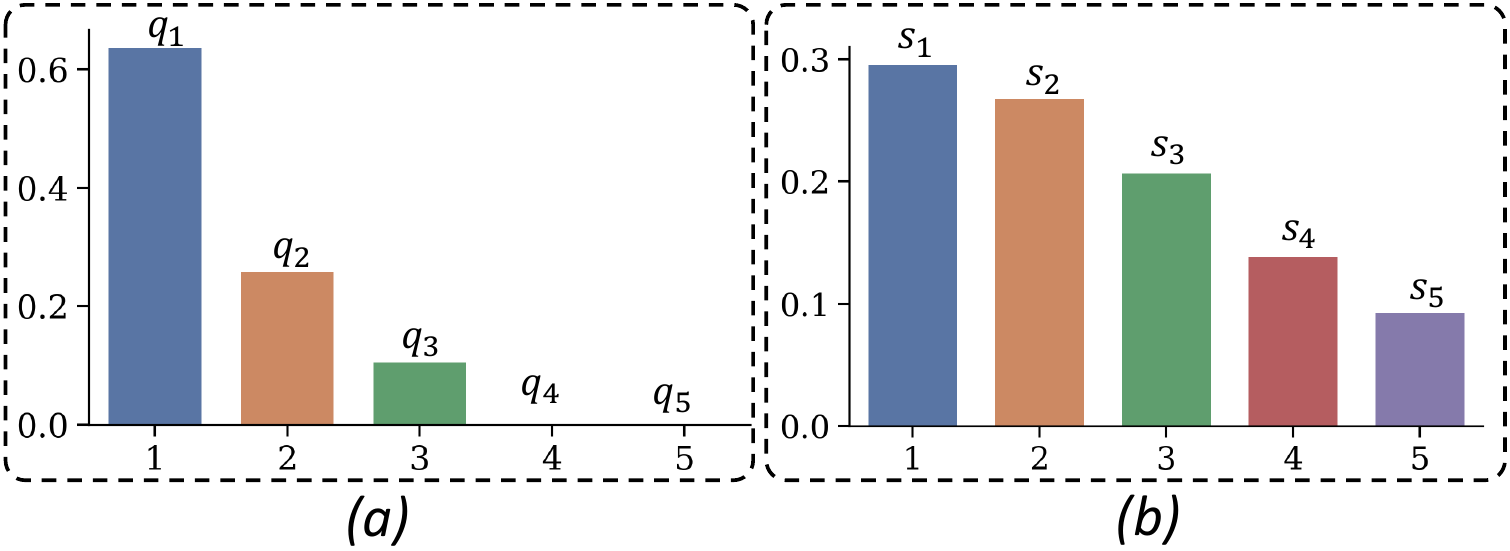}
    \caption{(a) KL-divergence-based distillation targets from cross-encoder. (b) Predicted similarity scores from student dual-encoder after softmax operation.}
    \label{fig:kl_distillation}
\end{figure}

Score distribution distillation (\ie, KL-divergence-based knowledge distillation) requires the student and teacher models have the same score distribution over multiple samples. Upon further analysis, we find that score distribution can be interpreted as ranking distillation with additional constraints. As shown in Figure \ref{fig:kl_distillation}, given an image and multiple texts $t_i,\ i\in \{1,2,\cdots,5\}$, we compute their similarity $p_i$ with cross-encoder and construct distillation target $q_i$ by applying softmax operation over these scores. A hyper-parameter $\tau$ is employed to control the sharpness of distillation target. Without loss of generality, we assume that:
\begin{equation}
    p_1 > p_2 > p_3 > p_4 > p_5. \label{eq1}
\end{equation}

We can prove that:
\begin{align}
    if\ p_i - p_j &> p_m - p_n, \notag \\
    then\ q_i - q_j &> q_m - q_n, \notag \\
    \forall i,j,m,n \in \{1,2,\cdots,5\},&\ i<j\leq m<n,\ \tau > 0. \label{eq2}
\end{align}

\textit{Proof.} According to Mean value theorem,
\begin{equation}
    q_i - q_j = \frac{e^{p_i/\tau} - e^{p_j/\tau}}{\sum_{k}e^{p_k/\tau}} = \frac{(e^a)'(p_i/\tau - p_j/\tau)}{\sum_{k}e^{p_k/\tau}},
\end{equation}
where $a \in (p_i/\tau, p_j/\tau)$. Similarly,
\begin{equation}
    q_m - q_n = \frac{(e^b)'(p_m/\tau - p_n/\tau)}{\sum_{k}e^{p_k/\tau}},
\end{equation}
where $b \in (p_m/\tau, p_n/\tau)$. Given the assumption of Equation \ref{eq1} and \ref{eq2}, we can derive that $a > b$ and thus $q_i - q_j > q_m - q_n$.

In other words, taking $t_1, t_2, t_3$ as examples, $p_1>p_2>p_3$ and $p_1 - p_2 > p_2 - p_3$, then the values $q_i$ satisfy $q_1 - q_2 > q_2 - q_3$ with any $\tau > 0$. Such a distillation target requires that:
\begin{equation}
    s_1>s_2>s_3, \label{eq5}
\end{equation}
\begin{equation}
    s_1 - s_2 > s_2 - s_3, \label{eq6}
\end{equation}
where $s_1,s_2,s_3$ is the similarity scores (after softmax) from student model. Note that the objective of Equation \ref{eq5} is the same as ranking distillation. However, the additional constraint of Equation \ref{eq6} may interfere with the learning of image-text alignment due to the significant difference between the similarity distributions of dual-encoder and cross-encoder, which is validated by our experimental results.

\section{Datasets Details}
\begin{table}[h]
 \caption{Statics of the pre-training datasets.}
 \label{tab:data_statistics}
 \centering
 \resizebox{0.8\linewidth}{!}{
 \begin{tabular}{lcccc}
    \toprule
     & COCO (Karpathy-train) & VG & CC3M & SBU\\
    \midrule
    image & 113K & 100K & 2.81M & 825K \\
    text & 567K & 769K & 2.81M & 825K \\
 \bottomrule
\end{tabular}}
\end{table}

\begin{figure*}[t]
    \centering
    \includegraphics[width=0.9\linewidth]{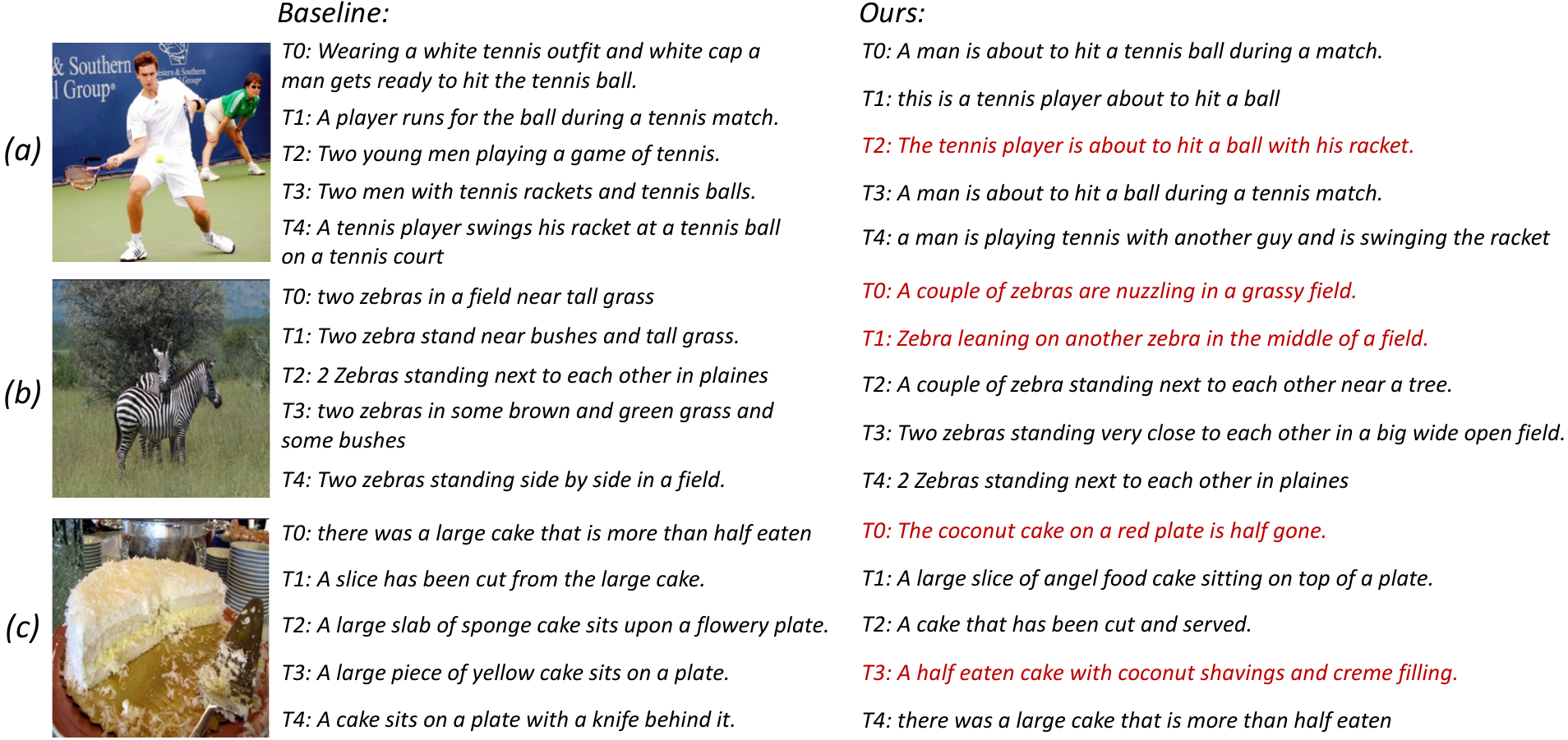}
    \caption{Illustration of image-to-text retrieval of our model and baseline model. Ground-truth captions for each image are in red color.}
    \label{fig:i2t}
\end{figure*}

\begin{figure}[t]
    \centering
    \includegraphics[width=0.9\linewidth]{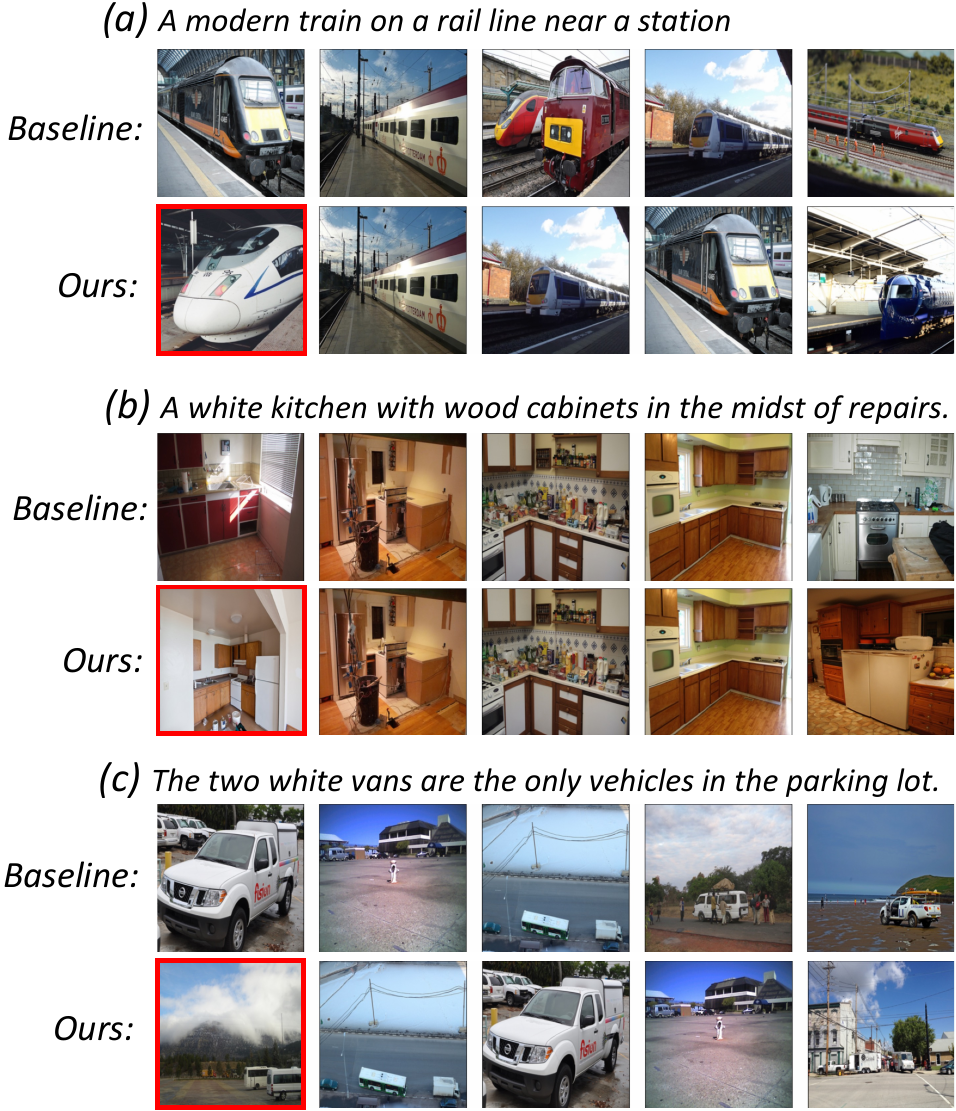}
    \caption{Illustration of text-to-image retrieval results of our model and baseline model. The ground-truth image for each text is in the red box.}
    \label{fig:t2i}
\end{figure}

\noindent \textbf{Pre-training datasets.} We show the statistics of the images and texts of pre-training datasets in the Table \ref{tab:data_statistics}

\noindent \textbf{MSCOCO.} MSCOCO \cite{lin2014microsoft} is a large image-text dataset of 123K images, where each image has 5 human-annotated captions. Following \cite{li2021align, kim2021vilt, lu2022cots}, we adopt the Karpathy split of MSCOCO, where 5K/5K/113K images are used for testing, validation and training respectively.

\noindent \textbf{Flickr30K.} Flickr30K contains 31K images and 159K captions. Each image is usually annotated with 5 captions. Following \cite{frome2013devise}, we 1K/1K/29K images for testing, validation and training respectively.

\noindent \textbf{Crisscrossed Captions.} Crisscrossed Captions dataset \cite{parekh2021crisscrossed} is an extension of MS-COCO dataset with human semantic similarity judgments for intra- and inter- modality pairs. It contains human ratings for 267,095 pairs (derived from 1,335,475 independent judgments), a massive extension in scale and detail to the 50k original binary pairings.

\section{Evaluation Metrics}
\noindent \textbf{Retrieval.} We report the widely-used R@\textit{k} (\textit{k}=1,5,10) for cross-modal retrieval, which is the proportion of matched samples found in the top-\textit{k} retrieved results. We also report R@S to reveal the overall performance, which is defined as the sum of R@\textit{k} metrics at \textit{k}=\{1,5,10\} of both image-to-text and text-to-image retrieval tasks.

\noindent \textbf{Ranking.} We report the Spearman’s bootstrap correlation following \cite{parekh2021crisscrossed, jia2021scaling} to assess whether a
model ranks pairs similarly to human raters.
For each correlation estimate,
we sample half of the queries (to increase diversity across samples) and for each selected query, we
choose one of the items for which Crisscross caption dataset supplies a
paired rating. We compute Spearman’s correlation between the ground-truth scores and the model scores for the selected pairs. The final correlation is the average
over 1000 of these bootstrap samples.

\section{Visualizations}

\noindent \textbf{Image-to-text Retrieval.} We show image-to-text retrieval results on the MSCOCO test set in the Figure \ref{fig:i2t}. We can observe that: (1) Our model has a more precise perception of detailed objects and actions in the image, \eg, the baseline model erroneously identifies "white cap", "run" from the (a), while our method accurately determines that it is a man hitting a ball with a racket; (2) Our model correctly recognizes detailed relation ``nuzzling" and ``leaning" in the (b), while the baseline model fails to achieve such recognition; (3) Our model achieves better cross-modal matching for rare concepts, as shown in (c), where our model recognizes the ``coconut" and aligns it with the corresponding text.

\noindent \textbf{Text-to-image Retrieval.} The text-to-image results are shown in Figure \ref{fig:t2i}. It can be seen that: (1) Our model perceives abstract adjectives more accurately, \eg, ``a modern train" in (a); (2) Our model understands local text semantics ``in the midst of repairs" better and find the image that contains repair tools in (b), but the baseline model only finds the images with ``kitchen" and ``cabinets"; (3) Our model has better understanding on the number, \eg, our model find the image with only ``two" white vans accurately in (c).

\section{Ablation Study}

\begin{table}[h]
 \caption{The Spearman's rank correlation ($\times$100) of samples from different ranking intervals between DE and CE.}
 \label{tab:spearman}
 \vspace{-0.3cm}
 \footnotesize
 \centering
 \resizebox{0.8\linewidth}{!}{
 \begin{tabular}{lcccc}
    \toprule
    \multirow{2}{*}{Rank Interval} & \multicolumn{2}{c}{image$\rightarrow$text} & \multicolumn{2}{c}{text$\rightarrow$image}\\
     & DE & +CPRD & DE & +CPRD \\
    \midrule
    1-16 & 53.1 & 61.3 & 50.7 & 60.0 \\
    17-32 & 17.0 & 22.8 & 16.8 & 21.7 \\
    33-48 & 10.1 & 14.7 & 15.7 & 12.8 \\
    49-64 & 7.1 & 10.0 & 23.1 & 27.4 \\
\bottomrule
\end{tabular}}
\vspace{-0.3cm}
\end{table}

\noindent \textbf{The effect of ranking mimicking.} To validate whether our method mimics the ranking of cross-encoder, we use dual-encoder to retrieve the top 64 texts/images given each image/text of MSCOCO test dataset. Then we re-rank the retrieved texts/images in the different rank interval (\ie, 1-16, 17-32, 33-48, 49-64) with cross-encoder and compute the spearman's rank correlation. As shown in Table \ref{tab:spearman}, applying our CPRD method on the dual-encoder improves the rank correlation on most of the rank intervals, validating the effectiveness of our method in mimicking cross-encoder's ranking. It is worth noting that the rank correlation degrades for top 33-48 retrieved images given texts, but the relative order between these lower-ranked samples is not important and our method is designed to disregard this order.

\begin{table}[h]
 \caption{The performance comparison with variation of $\mathcal{L}_{ij}$.}
 \label{tab:comparison_loss_var}
 \vspace{-0.25cm}
 \footnotesize
 \resizebox{\linewidth}{!}{
 \begin{tabular}{lccccccc}
    \toprule
    \multirow{2}{*}{Loss Type} & \multicolumn{3}{c}{image$\rightarrow$text} & \multicolumn{3}{c}{text$\rightarrow$image} & \multirow{2}{*}{R@S}\\
     & R@1 & R@5 & R@10 & R@1 & R@5 & R@10 & \\
    \midrule
    None & 32.0 & 59.4 & 71.5 & 24.4 & 49.5 & 61.0 & 297.8 \\
    $\hat{\mathcal{L}}_{ij}$ & 31.3 & 59.7 & 71.1 & 23.9 & 48.1 & 59.5 & 293.6 \\
    $\mathcal{L}_{ij}$ & \textbf{34.3} & \textbf{61.4} & \textbf{73.2} & \textbf{27.0} & \textbf{52.8} & \textbf{64.5} & \textbf{313.2} \\
\bottomrule
\end{tabular}}
\vspace{-0.3cm}
\end{table}

\noindent \textbf{The variant of our proposed contrastive partial ranking distillation loss.} 
Here, we want to explore ``Does it important to constrain that valid hard negatives have higher score than easy negatives in our proposed loss?". Without such constraint, the scores of hard negatives ranked lower are trained to have smaller similarity with CPRD, and might even be lower than those easy negatives, which have a negative impact on the performance of the dual-encoder. We test the variant loss $\hat{\mathcal{L}}_{ij}$ which does not have the above constraint. The original $\mathcal{L}_{ij}$ and $\hat{\mathcal{L}}_{ij}$ are formulated as:
\begin{align}
    &\mathcal{L}_{ij} = -log \frac{\mathrm{exp}(\bm{v}^\top_i\bm{\hat{t}}_{c_{ij}} / \tau)}{\sum \limits_{k=j}^{\mathrm{K}} \mathrm{exp}(\bm{v}^\top_i\bm{\hat{t}}_{c_{ik}} / \tau) + \sum \limits_{k=\mathrm{K}+1}^{B+N_q-1} \mathrm{exp}(\bm{v}^\top_i\bm{\hat{t}}_{d_{ik}} / \tau)}. \notag \\
    &\hat{\mathcal{L}}_{ij} = -log \frac{\mathrm{exp}(\bm{v}^\top_i\bm{\hat{t}}_{c_{ij}} / \tau)}{\sum \limits_{k=j}^{\mathrm{K}} \mathrm{exp}(\bm{v}^\top_i\bm{\hat{t}}_{c_{ik}} / \tau)}. \notag
\end{align}

As shown in Table \ref{tab:comparison_loss_var}, $\hat{\mathcal{L}}_{ij}$ is not as good as $\mathcal{L}_{ij}$, and it even has a negative impact on the baseline model, validating the importance of ensuring that valid hard negatives have higher score than easy negatives in the distillation loss.

\textbf{The choices between online hard negatives similarity calculation and offline approach.} As mentioned in Sec 3.2.2, using the cross-encoder to calculate similarity scores online brings additional training costs. To reduce the training cost, we can calculate the similarity of hard negative pairs in an offline manner. It is worth noting that, compared to online method, the offline computation for one teacher is heavier due to larger candidate number but only occurs once. Offline method is thus more efficient when reusing ranking targets (\eg, training multiple students with one teacher). Otherwise (\eg, training a student with varying teachers), online method is more efficient. The method choice depends on the scenarios.

\end{document}